\documentclass[letterpaper, 10 pt, conference]{ieeeconf}  

\IEEEoverridecommandlockouts                              

\usepackage{graphicx}
\usepackage{url}

\title{\LARGE \bf
Real-Time Service Robot Replanning via Simple Button Interaction \\for Improved Task Success and User Experience
}

\author{Ryo Terashima$^{*}$$^{1}$, Yuga Yano$^{*}$$^{1}$, Koshun Arimura$^{1}$ and Hakaru Tamukoh$^{1, 2}$
\thanks{This paper is based on results obtained from project JPNP16007 commissioned by the New Energy and Industrial Technology Development Organization (NEDO). This work was supported by JSPS KAKENHI Grant Numbers 23H03468 and 23K18495. This work was supported by JST ALCA-Next Grant Number JPMJAN23F3. This work was supported by JST SPRING, Japan Grant Number JPMJSP2154.}
\thanks{We would like to thank all the Hibikino-Musashi@Home members who helped us with the experiments.}
\thanks{All authors are with Kyushu Institute of Technology, Fukuoka, Japan. $^{*}$co-first authors
        {\tt\small\{terashima.ryo631, yano.yuuga158, arimura.koshun523\}@mail.kyutech.jp} and {\tt\small tamukoh@brain.kyutech.ac.jp}
        }
\thanks{Hakaru Tamukoh is also affiliated with the Research Center for Neuromorphic AI Hardware, Fukuoka, Japan.}
}

\begin{document}

\maketitle
\thispagestyle{empty}
\pagestyle{empty}

\begin{abstract}
Service robots must respond to unexpected instructions in real-world environments.
However, robots cannot detect all failures and exceptions during a task. 
To address these issues, we propose a real-time feedback function that enables robots to modify their behavior based on human feedback.
In this system, users can intuitively send feedback to the robot by pressing a single button on a tablet when the robot fails to act correctly.
Robots use this feedback to consider their failures and replan appropriate actions to complete the task.
We conducted experiments with and without the feedback function to verify the following hypothesis: ``Simple interactions do not cause a negative user experience.''
All questionnaire responses are evaluated on a five-point Likert scale.
After adding the feedback function, the response score for the question ``Did you feel that the robot’s behavior was unexpected?'' improved by 0.5 points, and that for ``Did you feel anxious about the robot's behavior at times?'' improved by 0.9 points. 
These results support the study's hypothesis and indicate that incorporating this real-time feedback function can simultaneously improve task success and the user experience.
\end{abstract}


\section{Introduction}
In recent years, service robots have been widely used in various contexts, such as restaurants, hospitals, and homes~\cite{Ono2022, Waitersystem2023, holland2021service, acar2024restaurant, yamaSII}. 
Service robots in these real-world environments often receive unexpected instructions.
Therefore, service robots need to understand instructions and respond flexibly according to the situation.

To achieve this goal, task planning methods based on large language models (LLMs) \cite{obinata2023foundation, Wang2024GPSR, memmesheimer2024EGPSR, yamao_icam} have been proposed for robots.
SayCan~\cite{SayCan} proposes a framework in which an LLM understands natural-language instructions and determines appropriate actions for the situation.
SayCan has an 87\% success rate for tasks requiring a single action.
However, this success rate decreases to less than 50\% for tasks requiring multiple actions.
Robots must detect failures and replan tasks to improve the success rates of complex tasks.
To this end, task replanning methods~\cite{KNOWNO, Self-recovery} have been proposed for service robots based on uncertainty prediction and self-recovery prompting.
Using these methods, when robots notice a problem during a task, they select an appropriate action by asking humans for help.
Therefore, the robots must detect failures in order to request help.
However, robots cannot detect certain failures, such as object or voice misrecognition.
Such situations necessitate a function that enables human intervention to correct the robot's behavior.

However, prior research indicates that increased interaction is closely linked to user burden. 
For example, increased interaction time reduces the overall efficiency of human-robot teams~\cite{olsen2003metrics}.
Additionally, user intervention during tasks decreases user performance and increases user frustration and anxiety~\cite {bailey2006need, mannem2023exploring}.
In fact, some recent studies have adopted designs that minimize human-robot interaction ~\cite{KNOWNO, gucsi2025user}.

Nonetheless, robots cannot autonomously detect all of their failures.
Thus, a gap exists between the design approach of ``intervention with robots is a burden to users and should be minimized'' and the reality that ``robots cannot achieve perfect autonomy.''

Prior studies~\cite{olsen2003metrics, bailey2006need, mannem2023exploring} indicate that the level of burden depends not only on interaction frequency but also on the interaction method.
Therefore, this study addresses the following hypothesis:
\noindent
\textbf{Hypothesis}: ``Simple interactions do not cause a negative user experience.''
In this study, we consider the problem to be not the intervention itself, but rather the burden the intervention design imposes on the user.

\begin{figure}[tb]
   \centering
   \includegraphics[width=\linewidth]{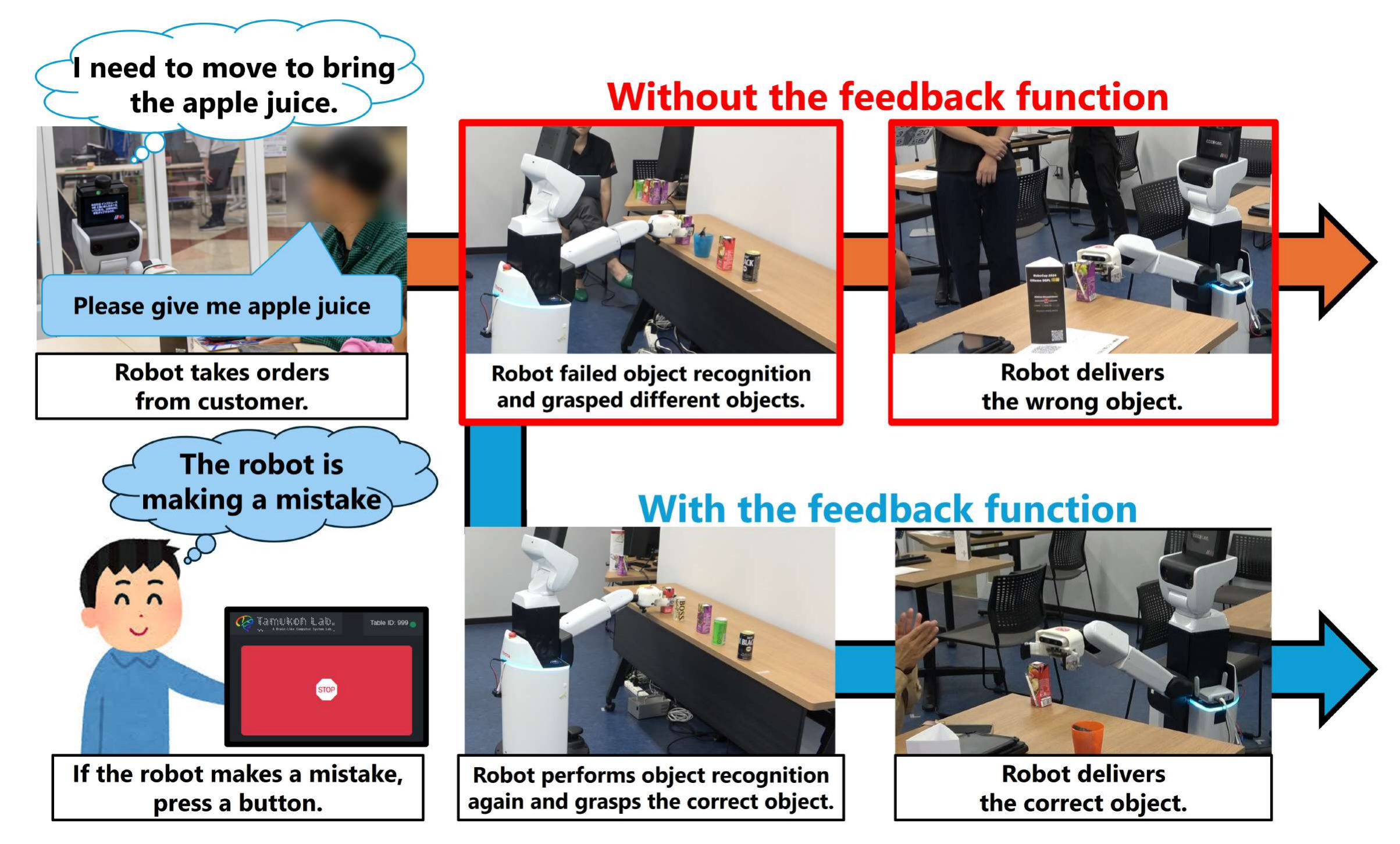} 
   \caption{Two scenarios in which a service robot delivers objects to a customer, with and without the proposed method. \textbf{Top}: Without the feedback function, the robot misrecognizes the object and delivers the wrong item. \textbf{Bottom}: With the feedback function, the user presses the ``stop'' button in case of failure, prompting the robot to reconfirm the object and deliver the correct item.}
   \label{Feedback_System}
\end{figure}

Therefore, we propose a real-time feedback system that allows users to easily intervene with a single operation when the robot fails at a task, thereby seamlessly modifying the robot's actions.
The simplicity of this interface is enabled by delegating failure reasoning to the LLM, thus eliminating the need for users to identify or articulate what went wrong.
Fig. \ref{Feedback_System} shows an overview of the proposed method.
The proposed method executes the sequential inference of tasks based on SayCan.
When the robot receives feedback during task execution, the robot replans its action using the situation when the failure occurred and the skills previously performed by the robot as input.
The feedback function enables robots to replan actions depending on the situation, improving task success rates.

We conducted experiments with and without the proposed feedback function in a simulated restaurant environment.
The experiment involved robots serving customers in a restaurant where orders change, and unexpected orders are placed.
After both experiments, we asked participants to complete an identical human-robot interaction (HRI) questionnaire.
We compared the task success rates and questionnaire results to verify the effectiveness of the proposed method.
The experimental results confirmed that implementing the real-time feedback system improved task success rates.
Furthermore, we confirmed that subjective evaluations indicated reduced user anxiety and a reduced desire to intervene with the robot.

The main contributions of this study are as follows:
\begin{itemize}
    \item We propose a real-time feedback system that enables users to easily intervene with a single operation, thereby modifying robot behavior during task execution.
    \item Through experiments and HRI questionnaires, we demonstrated that simplifying robot intervention operations reduces user burden.
\end{itemize}

\section{RELATED WORKS}
\subsection{LLM-Based Task Planning Methods}
In recent years, task planning methods based on LLMs have been proposed~\cite{SayCan, YanoROMAN2024}, enabling robots to interpret natural-language instructions and select situationally appropriate behaviors.
For example, SayCan~\cite{SayCan} proposed a mechanism in which an LLM combines predefined skills based on natural-language instructions to infer appropriate actions.
In addition, Yano et al. proposed a system~\cite{YanoROMAN2024} that integrates task understanding via LLM and response generation using dynamic maps for HRI in real-world environments.
However, while these frameworks achieve high success rates on single-step tasks, they exhibit greater uncertainty in multi-step tasks, leading to failures.
Here, a single-step task uses one skill (e.g., grasping), while a multi-step task chains several skills (e.g., grasping, moving, placing).

\subsection{Handling Failures in Robot Tasks}
\label{section:handling}
To address uncertainties and failures during robot task execution, approaches based on both prevention and recovery have been proposed~\cite{KNOWNO, Self-recovery}.
Ren et al. proposed KnowNo~\cite{KNOWNO}, which statistically organizes the uncertainty of LLM~\cite{anil2023palm2technicalreport} planners using conformal prediction and requests human assistance when uncertainty is high.
When the user provides a natural-language instruction, KnowNo generates multiple action candidates and constructs a prediction set that includes the correct action, meeting or exceeding a specified confidence level.
If the prediction set contains only a single candidate, the robot executes the action. If multiple candidates exist, the robot asks humans for help.
Shirasaka et al. proposed Self-Recovery Prompting~\cite{Self-recovery}, which enables robots to replan when they detect failures or insufficient information and to ask humans for help.
The system integrates several fundamental models, including plan generation with GPT-4 \cite{gpt4}, speech recognition with Whisper \cite{Whisper}, and object detection with Detic \cite{Detic}.

In these approaches, the robots themselves detect uncertainty or failure and request human assistance.
However, in real-world environments, robots cannot detect some failures, such as semantic misunderstandings or object recognition errors, on their own.
If robots cannot recognize their own failures, they cannot even ask humans for help.

\subsection{User Burden due to Interaction}
In situations where humans and systems collaborate on tasks, interactions are sometimes discussed in terms of user burden.
Olsen et al.~\cite{olsen2003metrics} formalized the amount of a user's attention consumed by interactions with a robot, demonstrating that attention resources constrain the efficiency of human-robot teams.

Additionally, user burden depends not only on the number of interactions but also on the timing of interventions.
Bailey et al.~\cite{bailey2006need} conducted a controlled experiment to evaluate the impact of the system's intervention timing on user burden.
Participants performed tasks on a computer and received interruptions from the system either at natural task breakpoints or during task execution.
The study found that interruptions during task execution significantly increased task completion time, error rate, discomfort, and anxiety compared to interruptions at breakpoints.
Furthermore, Mannem et al.~\cite{mannem2023exploring} investigated the cost of robot interventions from the user perspective in human-robot collaborative scenarios.
In this study, the timing, content, and proximity of questions posed by robots to human teammates during task execution varied.
All three factors were found to contribute to affect task performance.
Robots that intervened with teammates received lower usability ratings from users.
These findings suggest that robot-initiated inquiries negatively impact not only performance but also users' emotional evaluations of the robots.

Based on this background, recent robot research has often adopted a policy of minimizing user interaction~\cite {KNOWNO, gucsi2025user}.




\section{Proposal}
\begin{figure}[tb]
   \centering
   \includegraphics[width=0.48\textwidth]{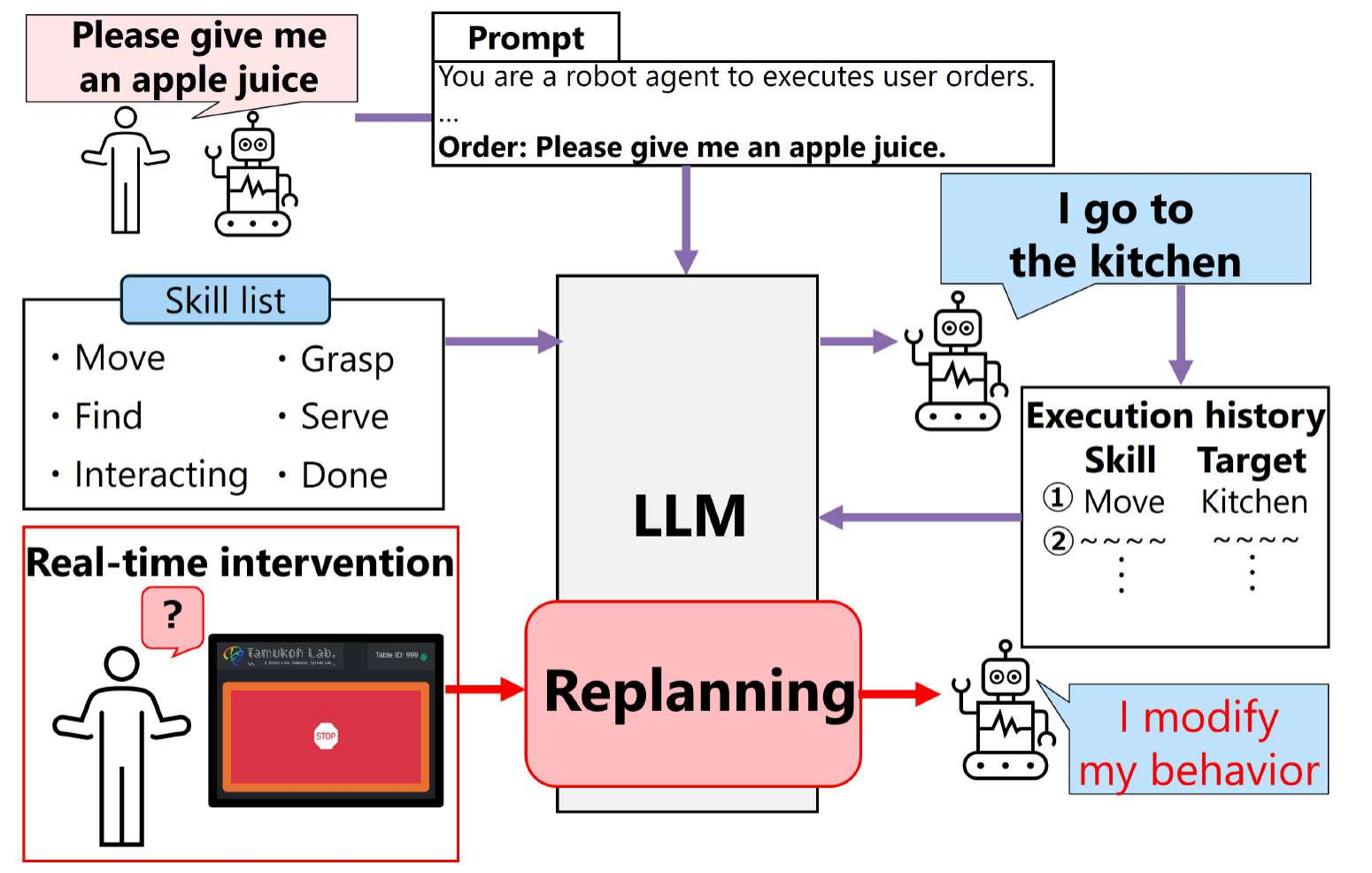} 
   \caption{Overview of the proposed system.}
   \label{proposed}
\end{figure}

This study aims to enable the robot to recover from failures that it cannot detect, while minimizing the intervention burden imposed on the user.
We propose a seamless real-time feedback function that enables humans to interact intuitively and easily with robots.
Fig. \ref{proposed} shows an overview of the proposed method.
The feedback function allows the robot to seamlessly modify its tasks by receiving feedback and modifying its behavior when the robot cannot detect failures itself.
We designed the feedback function to be simple, allowing users to send feedback with a single button press to minimize user interactions.

\subsection{Task Planning}
\begin{figure}[tb]
   \centering
   \includegraphics[width=0.48\textwidth]{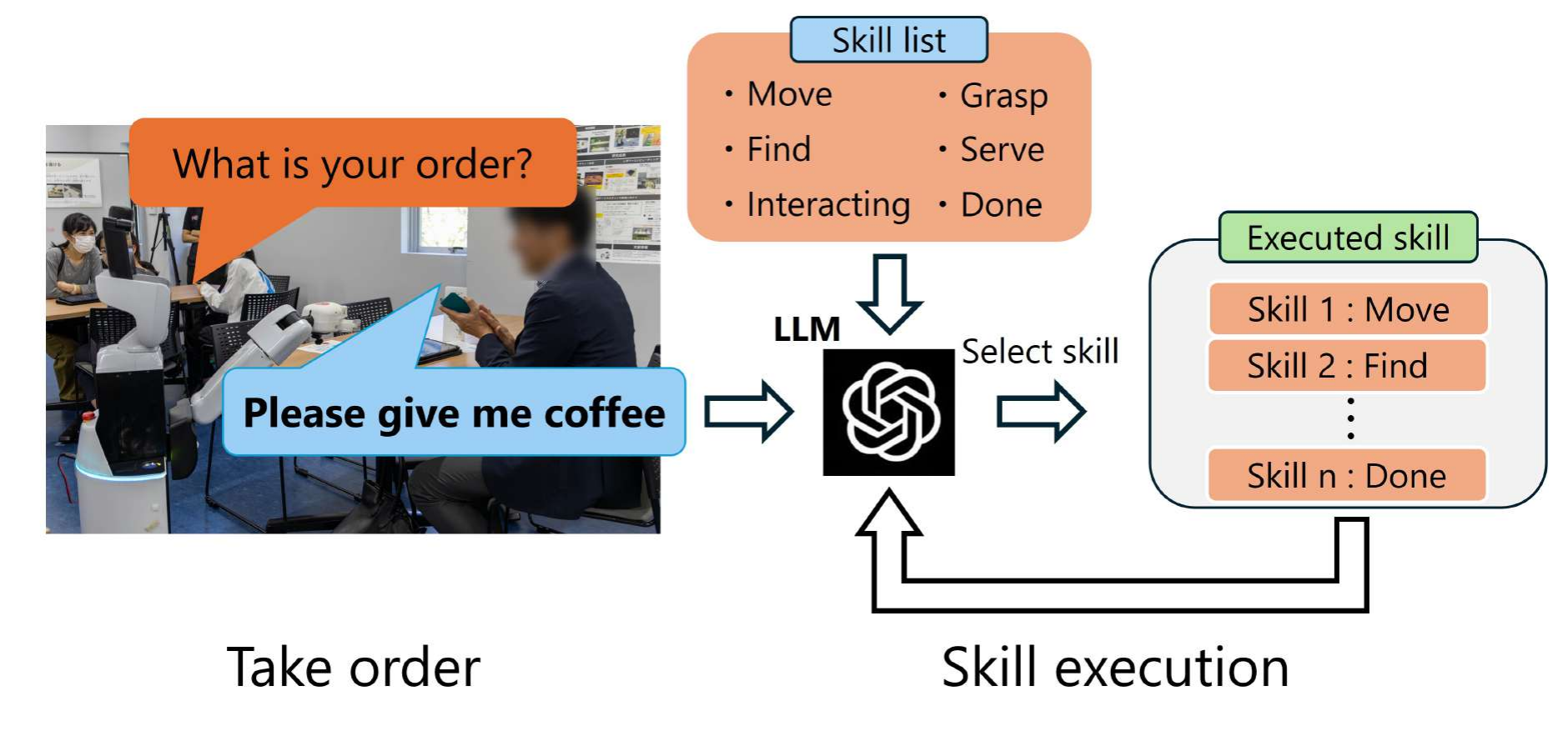} 
   \caption{The task planning system used in this study.}
   \label{system_flow}
\end{figure}

Fig. \ref{system_flow} shows a flow chart of the task planning method used in this study.
The proposed system can respond seamlessly to sudden modifications during tasks by performing sequential executions.
Yano et al.'s system executes skills by predicting the appropriate task from predefined tasks based on the user's instructions.
However, Yano et al.'s method cannot respond to failures during a task.
In this study, we improved the skill execution method and adopted sequential execution, which is based on SayCan \cite{SayCan}.
The proposed system inputs a list of skills and user instructions into GPT-4o \cite{gpt4o}, which first infers skills to be executed and determines the target(object or location).
After executing the skill, the system infers the next skill and target to execute by using the previously executed skills and the target object or location.
When there are no skills to execute, the result of the inference is ``Done'' and task planning is finished.

\subsection{Real-Time Feedback Function}
\begin{figure*}[tb]
   \centering
   \includegraphics[width=\textwidth]{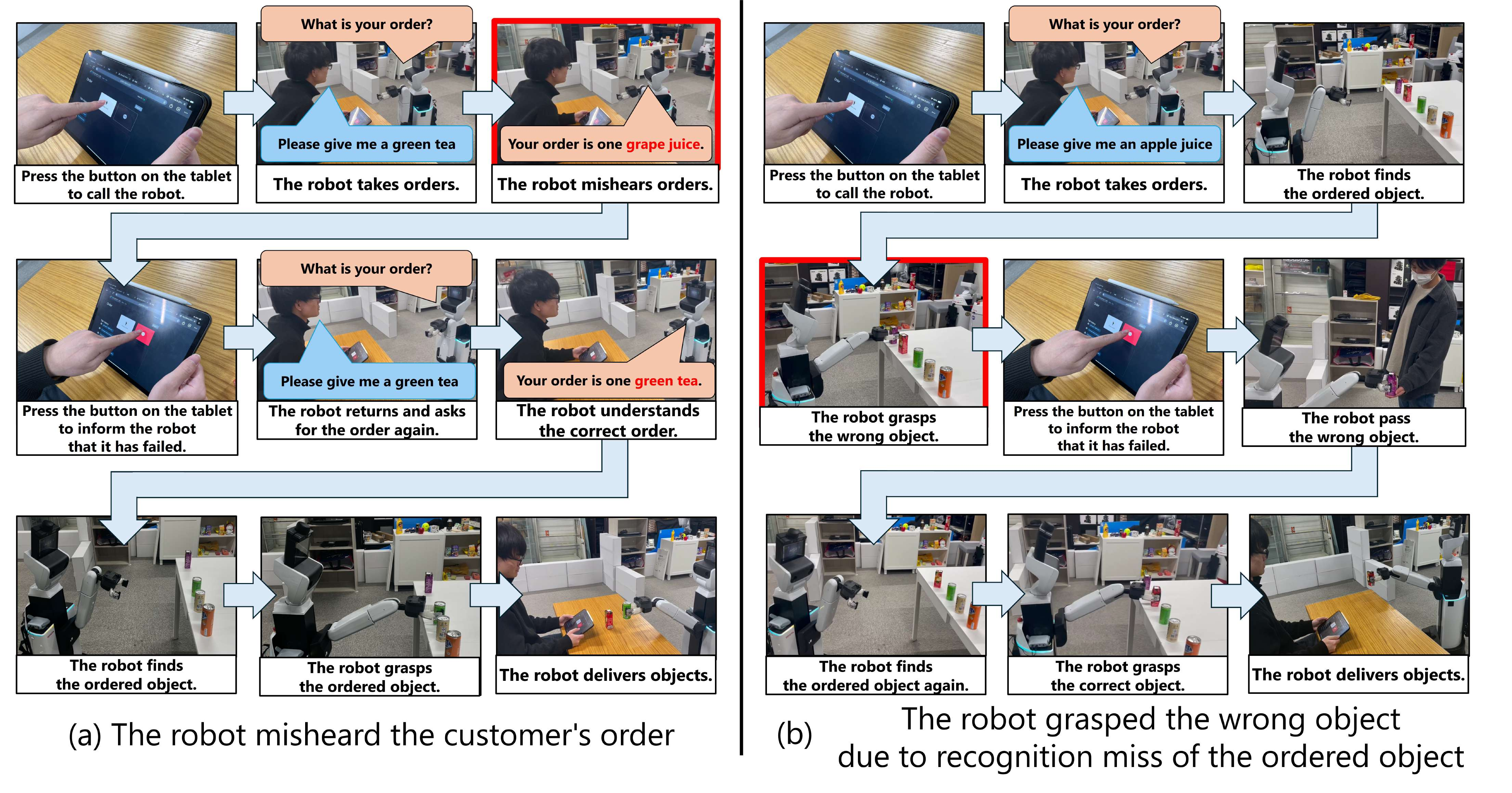} 
   \caption{Examples of the use of the feedback function.}
   \label{feedback_example}
\end{figure*}

\begin{figure}[tb]
   \centering
   \includegraphics[width=0.48\textwidth]{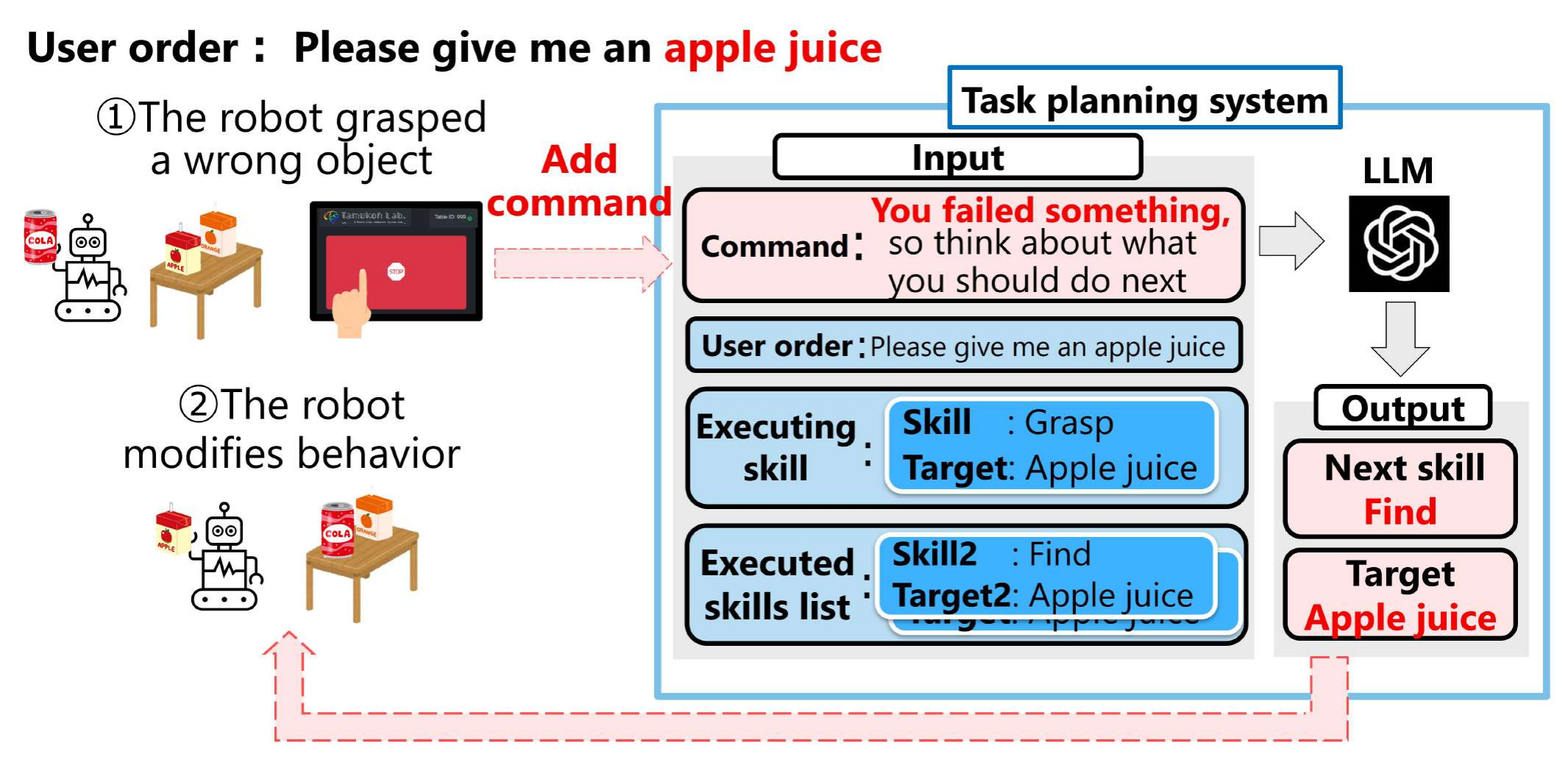} 
   \caption{Example of prompts sent to the proposed system when using the feedback function shown in Fig. 2.}
   \label{prompt_example}
\end{figure}

As a key design feature of the feedback function, failure cause analysis is assigned to the LLM.
Therefore, users do not need to identify the problem or explain the issue to the robot.
Consequently, we can simplify the user interface to a single button press, minimizing the physical and cognitive burden on the user.
Fig. \ref{feedback_example} shows examples of the feedback function.
Users can easily send feedback by simply pressing a button on the tablet.
This button only sends feedback to the robot that ``something failed,'' regardless of the task execution stage at which it is pressed.
Fig. \ref{prompt_example} shows examples of the prompt to send to the task planning system when using the feedback function.
Prompts include a list of executed skills and a user's order.
By including this information, the robot can understand its situation. 
After that, the robot infers ``What did I fail?'' and ``How should I modify my behavior?''.
The robot then reconsiders the target object and skills, such as moving and grasping, based on this inference.
For example, if the robot receives feedback when grasping the wrong drink, the robot will consider that the drink grasped was incorrect.
Then, the robot modifies the task to find and grasp the correct drink.

\begin{figure}[tb]
   \centering
   \includegraphics[width=\linewidth]{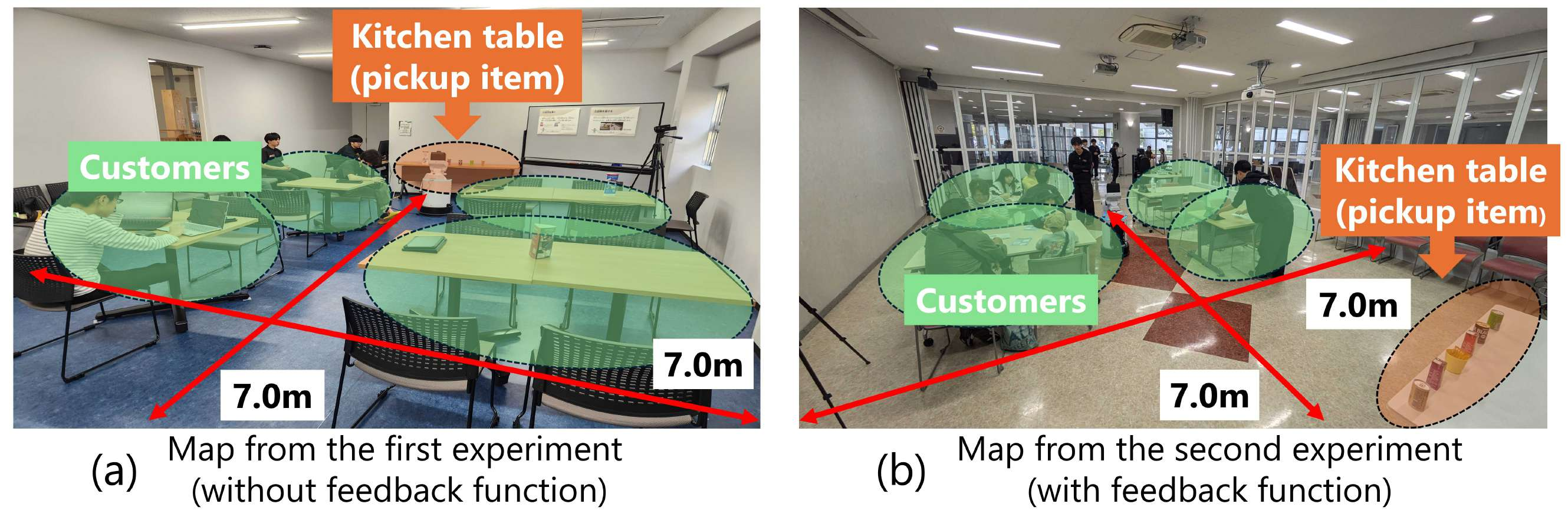} 
   \caption{Description of experimental field map.}
   \label{map}
\end{figure}

\section{Experiment}
Restaurants increasingly require service robots to help waiters with tasks owing to labor shortages.
Restaurant waiter tasks are challenging because robots require flexible responses to changes in orders, unexpected requests, and changes in real-world environments.
Therefore, we focused on waiter tasks, including understanding orders, serving food, cleaning tables, and interacting with users.
In the experiment, we designed waiter tasks based on RoboCup@Home \cite{robocup}, which is intended to operate a service robot indoors.
In this study, we conducted user experiments on a waiter task to verify the following hypothesis:
\textbf{Hypothesis}: ``Simple interactions do not cause a negative user experience.''
In this study, we regard a negative user experience as the user forming a bad impression of the robot, such as feeling anxious or perceiving its behavior as unexpected.

\begin{figure}[tb]
   \centering
   \includegraphics[width=0.48\textwidth]{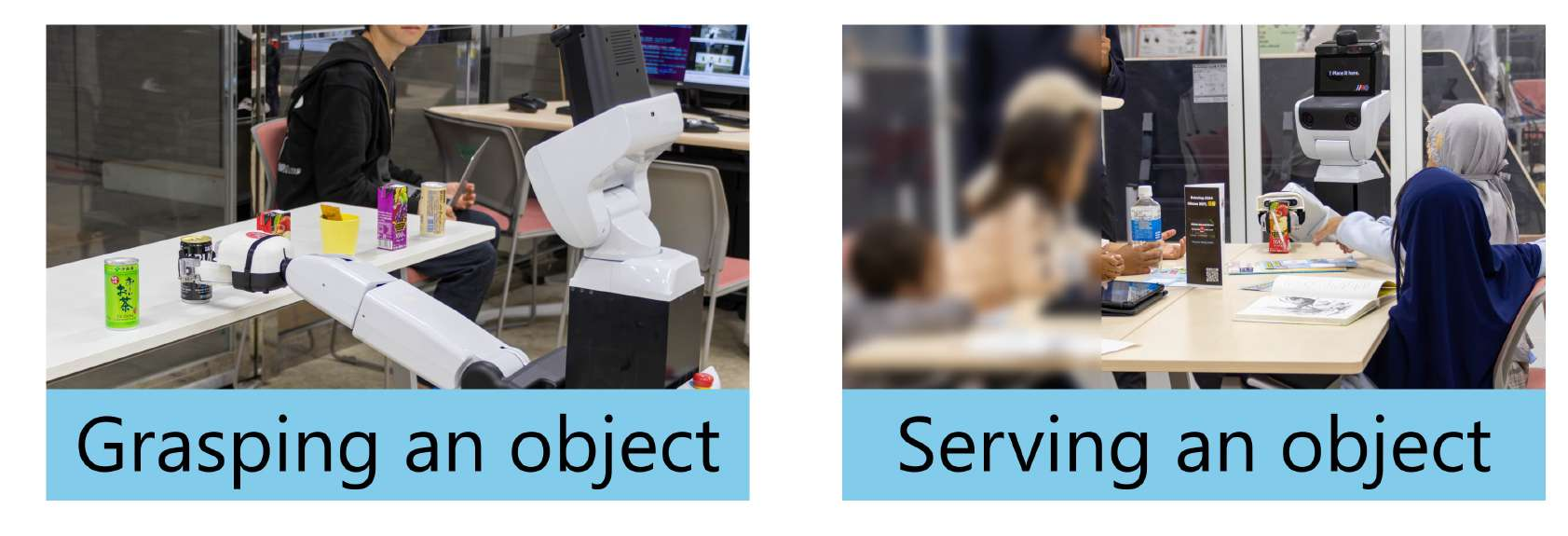} 
   \caption{The experimental scenario for evaluating the proposed method.}
   \label{experiment_example}
\end{figure}

To verify this hypothesis, we conducted experiments under the same conditions with and without the feedback function, comparing task success rates and questionnaire results.
The questionnaire is based on a five-point Likert scale~\cite{joshi2015likert}.
Fig. \ref{map} shows two experimental environments in this study.
Fig. \ref{experiment_example} shows the scenario of the actual experiment.
In the two experiments, we invited different participants under the same settings, including the field size, desk layout, and object.
Each experiment involved a group of participants, so the number of participants exceeds the number of trials. 
In these experiments, we used Human Support Robot \cite{yamamoto2019hsr}, a mobile manipulator equipped with a microphone and display that enables interaction with humans.

\subsection{Experimental Result Without the Feedback Function}
Table.~\ref{tab:task_success_rate} shows the task success rate of experiments.
We experimented with 91 users for which robots performed order-taking and serving tasks 38 times.
The robot successfully delivered the requested items 21 times out of 38.
The main reasons for these failures include errors in speech recognition, such as mishearing the order, and misrecognition of other objects as the target object.

\begin{figure}[tb]
   \centering
   \includegraphics[width=0.48\textwidth]{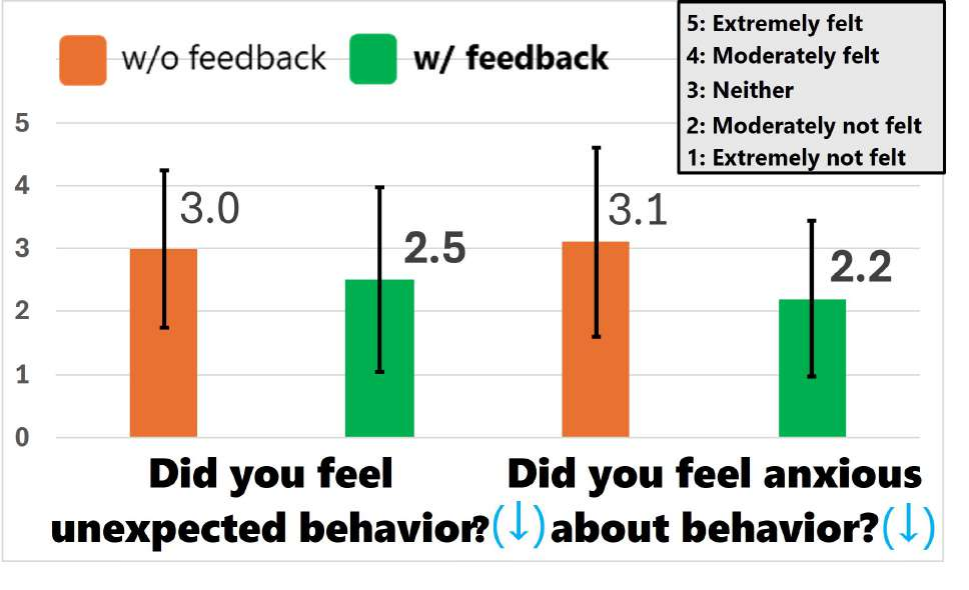} 
   \caption{Summary of questionnaire results from the experiment with and without feedback.}
   \label{questionnaire}
\end{figure}

Fig. \ref{questionnaire} shows the results of the questionnaire.
Thirty-six users responded to the questionnaire regarding the HRI.
Ten of the respondents were students, and 26 were working adults.
The question ``Did you feel that the robot’s behavior was unexpected?'' resulted in an average of 3.0 points overall.
These results indicate that the robot sometimes fails, and many users feel inconvenienced because users cannot provide additional instructions or point out failures.
Responses to the question ``Did you feel anxious about the robot's behavior at times?'' resulted in an average of 3.1 points overall. 
These results indicate that without the feedback function, users do not fully understand or feel comfortable with the robot's behaviors.

\begin{table}[tb]
\centering
\caption{Task Success Rates With and Without the Feedback Function}
\label{tab:task_success_rate}
\begin{tabular}{l c c}
\hline
\textbf{Condition} & \textbf{Successes / Total Trials} & \textbf{Success Rate} \\
\hline
w/o feedback function & 21 / 38 & 55\% \\
w/ feedback function & 43 / 50 & \textbf{86\%} \\
\hline
\end{tabular}
\end{table}
We conducted a questionnaire regarding the need for a function to provide additional instructions in experiments without the feedback function.
The results showed that the question ``Did you want to give the robot additional instructions or point out failures?'' resulted in an average of 4.5 points overall.
These questionnaire results indicate that usability decreases when the robot behaves unexpectedly, and the feedback function is crucial for solving these problems.

\subsection{Experimental Result With the Feedback Function}
We experimented with 133 users, and the robot performed order-taking and serving tasks 50 times.
Out of 50 tasks, the robot successfully completed 43 tasks, of which 8 were recovered through the feedback function. 
The remaining 7 tasks failed.
This result indicates that the robot can reconsider and correct tasks when feedback is sent.

Thirty-six users responded to a questionnaire regarding HRI, 
16 and 20 of whom were students and working adults, respectively.
Fig. \ref{questionnaire} shows the results of the questionnaire.
Responses to the question ``Did you feel that the robot’s behavior was unexpected?'' resulted in an average of 2.5 points overall. 
This score improved by 0.5 points in the first experiment, indicating that more respondents felt that the robot's behavior was as intended after the feedback function was introduced.
Responses to the question ``Did you feel anxious about the robot's behavior at times?'' resulted in an average of 2.2 points overall. 
This result improved by 0.9 points compared with the first experiment, indicating decreased anxiety concerning robots among the respondents. 

These results support the hypotheses of this study.
Despite the feedback function increasing interaction between users and the robot, users reported reduced anxiety and a greater desire for robot intervention.
These findings suggest that the user burden of interacting with robots depends on the design of the intervention mechanism and that a single-operation, user-led design can improve the user experience.
Furthermore, we confirmed that users of all ages, including children, could utilize the feedback function, demonstrating the accessibility of the one-button interface.


\section{Discussion}
The questionnaire results confirmed that although the introduction of the real-time feedback system increased interaction frequency with users, user anxiety decreased, and the desire for more robot intervention increased.
Users pressed the feedback button only 8 times out of 50 tasks; over 90\% of users did not press the button. 
Nevertheless, users' anxiety scores improved.
These results suggest that the mere existence of a means to modify the robot's behavior reduces users' anxiety.
Therefore, we consider that interaction designs contribute to the user experience independently of task outcomes.
The findings support a key hypothesis of this study: ``Simple interactions do not cause a negative user experience.''

The implementation of the proposed system
improved task success rates from 55\% to 86\%.
The results indicate that user-initiated interventions are effective for failures that the robot cannot detect itself, demonstrating the system's ability to address the limitations discussed in Section~\ref{section:handling}.

Regarding differences between age groups, the adult group exhibited lower anxiety scores than the elementary school group.
One possible reason for this is that the working adults accompanying children focused more on safe, predictable robot behavior, thereby increasing their sensitivity to unexpected actions.
However, improvements observed in both groups through the feedback function suggest the proposed design is effective for a broad user base.

Some users observed a tendency not to use the feedback function even when the robot failed.
The tendency was especially evident among elementary school children.
The average score for the question ``Could you easily understand the robot's next action?'' was 3.5 points for elementary school children and 4.1 points for adults.
The results suggest that younger users found it difficult to predict the robot's next action and struggled to determine the appropriate timing for intervention.

In the experiment with the feedback system, in some cases, users did not notice the robot's failure and sent no feedback.
This case suggests a limitation of the current design: the proposed system relies on users to detect failures, making failure detection difficult when users cannot monitor the robot.
Note that in this experiment, participants observed the robot closely in an event setting. In more realistic situations where users do not constantly watch the robot, fewer failures may be caught, so the success rate could be lower.
The key point is that if the robot's current state is clearly indicated, users do not need to monitor robots constantly.
An interface that enables users to instantly understand the robot's behavior can help users judge the need for intervention without requiring constant monitoring.
Designing an interface that allows users to instantly understand the robot's state, regardless of when users check, is an important future research topic.

In this study, we showed that a simple, single-operation interaction design reduces user burden. 
However, button use is only one example of intervention design; we need to investigate whether optimal methods exist.
Candidates include voice, gestures, and gaze detection.
These candidates involve trade-offs in usability, accessibility, and implementation complexity.

\section{Conclusion}
In this study, we proposed a real-time feedback function that enables service robots to flexibly modify their behaviors.
In the experiment, we compared the questionnaire results and success rates of waiter tasks in restaurants with and without the proposed feedback function.
The experimental results showed that the feedback function improved the success rate of serving food from 55\% to 86\%.
Additionally, the questionnaire results indicate that the feedback function made the robot's behavior easier to understand and reduced anxiety. 
The proposed simple and intuitive feedback function increased usability by providing users with more choices and enabling them to interact with robots more freely.
Therefore, user impressions were positive regarding HRI despite increased user interactions with the robot.
User impressions of HRI were positive, mainly because the proposed feedback function was simple to use and easily accessible to users of all ages.
These results support the study's hypothesis: ``Simple interactions do not cause a negative user experience.''

In future works, we will improve the methods for expressing information concerning the robot to help users more easily understand the robot's behavior.
In addition, we will consider implementing other user feedback methods, such as voice and gestures.
Finally, we will conduct user experiments in more fluid and complex environments, such as home environments.

\addtolength{\textheight}{-12cm}   

\bibliographystyle{ieeetr}
\bibliography{ref/ref.bib}

\end{document}